\documentclass{article}
\usepackage{spconf,amsmath,graphicx}
\usepackage{epstopdf}
\usepackage{array}
\usepackage{graphicx}
\usepackage{multirow}
\usepackage{color, soul}
\usepackage{cite}
\usepackage{url}
\usepackage{hyperref}

\title{Teacher-Student Training for Robust Tacotron-based TTS}

\name{Rui Liu$^{ 1,2}$ \thanks{\textbf{Speech samples:} https://ttslr.github.io/ICASSP2020/ }, Berrak Sisman$^{ 2, 4}$, Jingdong Li$^{ 1}$, Feilong Bao$^{ *1}$,  Guanglai Gao$^{ 1}$, Haizhou Li$^{ 2, 3}$}
\address{ $^1$ Inner Mongolia University, China   $^2$ National University of Singapore, Singapore\\  $^3$ University of Bremen, Germany  $^4$ Singapore University of Technology and Design (SUTD) \\
\small{ \{r.liu,berraksisman\}@u.nus.edu}, haizhou.li@nus.edu.sg
}

\begin{document}
\maketitle
\begin{abstract}
While neural end-to-end text-to-speech (TTS) is superior to conventional statistical methods in many ways, the exposure bias problem in the autoregressive models remains an issue to be resolved. The exposure bias problem arises from the mismatch between the training and inference process, that results in unpredictable performance for out-of-domain test data at run-time. To overcome this, we propose a teacher-student training scheme for Tacotron-based TTS by introducing a distillation loss function in addition to the feature loss function. We first train a Tacotron2-based TTS model by always providing natural speech frames to the decoder, that serves as a teacher model. We then train another Tacotron2-based model as a student model, of which the decoder takes the predicted speech frames as input, similar to how the decoder works during run-time inference. With the distillation loss, the student model learns the output probabilities from the teacher model, that is called knowledge distillation. Experiments show that our proposed training scheme consistently improves the voice quality for out-of-domain test data both in Chinese and English systems.

\end{abstract}
%


\vspace{-3mm}
\section{Introduction}
\label{sec:intro}
With the advent of deep learning, end-to-end  TTS has shown many advantages over the conventional TTS techniques \cite{tokuda2013speech, ze2013statistical,liu2017mongolian}. For example, Tacotron-based approaches \cite{wang2017tacotron,shen2018natural,liu2020wavetts,lee2019robust} with an encoder-decoder architecture and attention mechanism have shown to achieve remarkable performance. In these techniques, the key idea is to integrate the conventional TTS pipeline into a unified network and learn the mapping directly from the $<$text, wav$>$ pair \cite{lee2019robust,chung2019semi,He2019,Luong2019}. Furthermore, together with a neural vocoder \cite{hayashi2017investigation,shen2018natural,Okamoto2019, berrak_is18, berrak-journal, sisman2018adaptive}, natural-sounding human-like speech can be generated.

However, neural end-to-end TTS is still far from perfect. A typical neural TTS system suffers from the exposure bias problem \cite{ranzato2015sequence,schmidt2019generalization} in the autoregressive model \cite{juang1985mixture} that is used by the decoder module. Specifically, in training stage, the decoder generates a frame using its previous frames of natural speech as input, that is called \textit{teacher forcing mode}. However, in inference stage, the decoder predicts a frame using its previously predicted frames as input, that is also called \textit{ free running mode}. There exists a mismatch between the natural speech frames and the predicted frames especially for out-of-domain test data, that leads to unpredictable outcomes, such as skipping, repeating words, incomplete synthesis and inappropriate prosody phrase breaks \cite{He2019,2018interspeech,ren2019fastspeech,zhu2019pre}.
 

The techniques to improve in-domain performance of neural TTS frameworks include attention mechanism  \cite{tachibana2018efficiently} and scheduled sampling \cite{ bengio2015scheduled,huszar2015not}. The use of scheduled sampling comes with negative effects that include misalignment between the natural speech frames and the predicted frames due to the fact that the temporal dependency of the acoustic sequence is disrupted. The techniques to improve out-of-domain performance include the GAN-based TTS framework \cite{guohh2019gantts} that introduces both real and generated data sequences in discriminator training, and more recently, stepwise monotonic attention for neural TTS \cite{He2019}. 

In this paper, we propose a novel training scheme, the teacher-student training scheme, for neural end-to-end TTS framework, that performs remarkably well for out-of-domain inference. In this scheme, a teacher model learns the text-speech mapping from training data in teacher forcing mode, while a student model learns from both the probability distribution of the teacher model and the same training data for teacher model in free running mode. The process of student learning from teacher model is called knowledge distillation, and its learning objective is called distillation loss.  

The main contributions of this paper are summarized as follows: 1) we propose a compact method for end-to-end TTS model; and 2) we propose a teacher-student training scheme for Tacotron-based TTS model. To our best knowledge, this is the first implementation of teacher-student training scheme for Tacotron2 based TTS framework. The proposed training scheme is validated with out-of-domain test data in Chinese and English TTS systems. 

This paper is organized as follows: In Section 2, we re-visit the Tacotron2-based TTS framework that serves as a baseline reference.  In Section 3, we study the proposed teacher-student training scheme. In Section 4, we report the evaluation results. We conclude the paper in Section 5. 

\section{Tacotron2 based TTS}
\label{sec:baseline}

In this paper, we adopt Tacotron2 \cite{shen2018natural} with scheduled sampling in the training stage, as a reference baseline. For rapid turn-around, we use Griffin-Lim \cite{griffin1984signal}  waveform reconstruction instead of WaveNet vocoder in this study. We note that the selection of waveform generation technique will not affect our judgment of the effectiveness of the proposed training scheme.

\begin{figure}
\vspace{-3mm}
  \centering
  \centerline{\includegraphics[width=9cm]{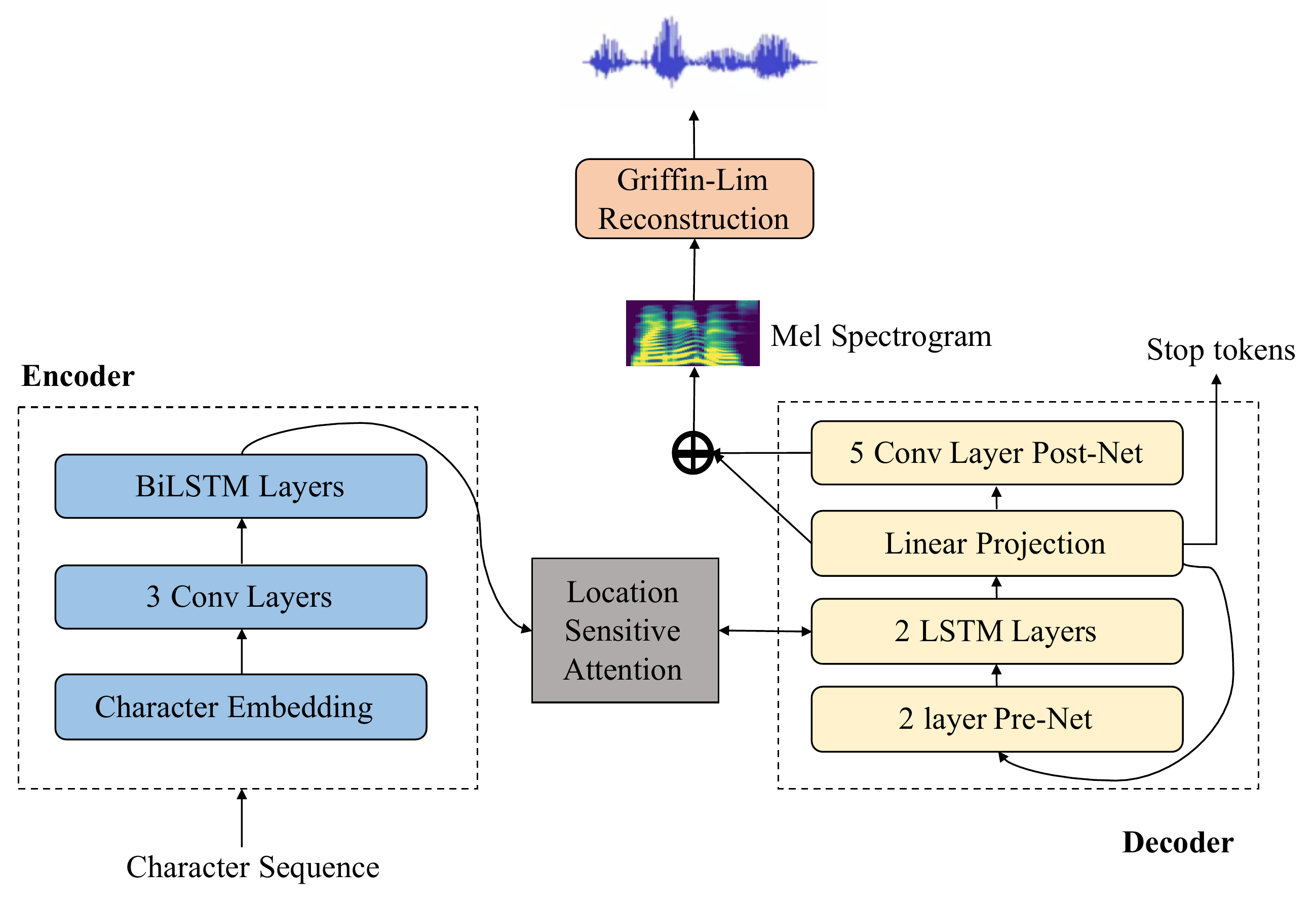}}
  \vspace{-5mm}
\caption{Block diagram of Tacotron2-based reference baseline that has three modules, encoder, attention-based decoder, and  Griffin-Lim reconstruction algorithm.}
\label{fig:baseline}
 \vspace{-5mm}
\end{figure}

We illustrate the overall architecture of the \textit{reference baseline} in Figure \ref{fig:baseline}, that includes encoder, attention-based decoder and Griffin-Lim algorithm. The encoder consists of two components, a CNN\cite{krizhevsky2012imagenet, ak2018learning} based module that has 3 convolution layers, and a LSTM\cite{emir2019semantically,zhang2018error} based module that has a bidirectional LSTM layer. The decoder consists of four components: a 2-layer pre-net, 2 LSTM layers, a linear projection layer and a 5-convolution-layer post-net. The decoder is a standard autoregressive recurrent neural network that generates the mel-spectrogram features and stop tokens frame by frame. 

During training, the decoder generates a frame in the scheduled sampling mode. However, at run-time inference, the decoder performs in free running mode to predict the future frames. Such trained decoder experiences the mismatch between the natural speech frames and the predicted speech frames, and the adverse effect of scheduled sampling on the temporal dependency of natural acoustic sequence. To address the above issues during training, we study a teacher-student training scheme in Section 3.


\section{Teacher-student training for Tacotron2 based TTS}

\label{sec:proposed}
\begin{figure*}[tb]
\vspace{-10mm}
\centering
\centerline{\includegraphics[width=17cm]{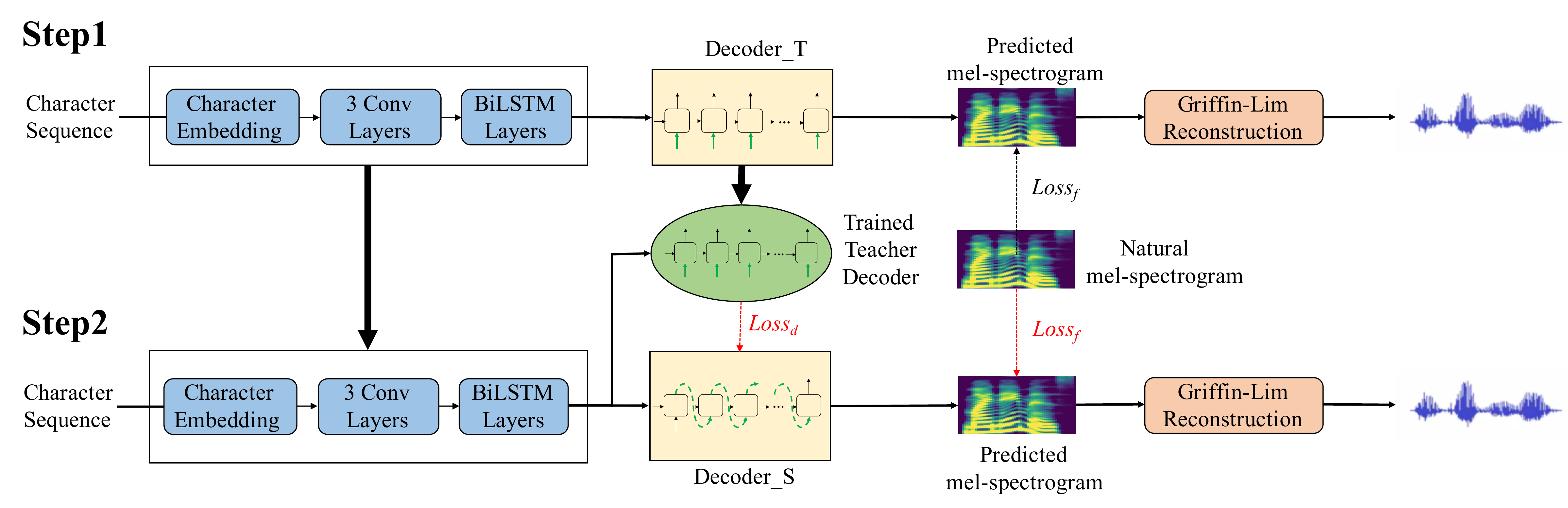}}
 \vspace{-5mm}
\caption{Illustration of the proposed teacher-student training scheme for Tacotron2-based TTS in 2 steps: Step 1, pre-train a teacher model, that includes teacher encoder and teacher decoder (``Decoder\_T''); Step 2, use the trained teacher encoder and teacher decoder to train the student decoder (``Decoder\_S'') by applying the proposed knowledge distillation approach.}
\label{fig:proposed}
\vspace{-3mm}
\end{figure*}

 In this section, we discuss in detail the teacher model, the student model, and the teacher-student training scheme. While both the teacher model and the student model have identical network architecture as the \textit{reference baseline }, they adopt different decoding strategies as illustrated in Figure \ref{fig:proposed}. 
 
 In practice, we first train a standard Tacotron2 teacher model for an end-to-end TTS system under the \textit{teacher forcing mode}, that is regarded as the teacher model. As the teacher model learns under the \textit{teacher forcing mode}, it is expected to represent the true distribution of the natural speech data. We then train another Tacotron2 student model under the \textit{free running mode}. The student model is trained by learning from both ground-truth sequence and the hidden states of the teacher model simultaneously. By learning from the hidden states of the teacher model via knowledge distillation, the student model learns the true distribution of the natural speech data effectively. As the student model is trained under the \textit{free running mode} by using the predicted speech frames as the input of the decoder, it is expected to accustom itself to the run-time inference condition. 

\subsection{Teacher Model}
\label{subsec:tehr}
For the decoder in teacher model, we implement the \textit{teacher forcing mode} that predicts a speech frame by taking the previous natural speech frames in the sequence as the input.  

Given a input character sequence $x = (x_{1}, x_{2}, ..., x_{T})$ and its target mel-spectrogram features $y = (y_{1}, y_{2}, ..., y_{T'})$, let $P(y|x,\theta)$ is the teacher model of which $\theta$ is the model parameters. Teacher model with teacher forcing mode takes the previous frames  $y_{1}, ..., y_{t-1}$ from the target natural speech as input to predict the feature frame $y_{t}$ at time step $t$, as formulated next, 

\begin{equation}
P(\hat{y}|x,\theta) = \prod _{t=1}^{T'} P(\hat{y}_{t}|y_{<t},x,\theta)
\end{equation}
where $\hat{y}$ is the predicted value and $y$ is from the target natural speech.

With such decoding mode, the teacher model is expected to learn the true probability distribution from natural speech data, that would be very informative for the student model.

\subsection{Student Model}
\label{subsec:stu}

The student model has the same network architecture as the teacher model, except that it has a completely different decoding mode: \textit{free running mode}.
In this mode, the decoder predicts  a  speech  frame  by  taking  the
previous predicted speech frames in the sequence as the input. The decoding process of the student model is defined as:

\begin{equation}
P(\hat{y}|x,\theta) = \prod _{t=1}^{T'} P(\hat{y}_{t}|\hat{y}_{<t},x,\theta)
\end{equation}
where $\hat{y}$ is the predicted value.

\subsection{Knowledge Distillation}
\label{subsec:loss}

Typically, knowledge distillation is a process where a small model is  trained to mimic a pre-trained, larger model \cite{yim2017gift}. In this paper, we borrow the idea of knowledge distillation in the implementation of the teacher-student training scheme.

The idea is to use a teacher model, that has been trained under the \textit{teacher forcing mode}, to guide the training of the student model, that runs under free running mode. As the teacher model is trained using natural speech frames as the input of decoder, we expect the output probability distribution of the teacher model to reflect the true distribution of the natural speech data. The student model is trained under the \textit{free running mode}. Therefore, it is closer to the actual inference condition.  At the same time, the hidden states of the student model are optimized to be close to those of the teacher model by way of knowledge distillation. As can be seen in Figure \ref{fig:proposed}, we define one objective function for the teacher model, the feature loss. We devise two objective functions for the student model, one for the feature loss that is the same as in the teacher model, and another for the knowledge distillation, or distillation loss. 

We formulate the entire  process  next. The encoder takes the input character sequence $x = (x_{1},x_{2},...,x_{T})$ from the given text and converts the one-hot vector to continuous high-level features representation $h$:
\begin{equation}
h_{t} = {\rm Encoder}(h_{t-1},x_{t})
\end{equation}

The teacher decoder, \textit {Decoder\_T} outputs a hidden state $s_t$ at each step $t$:
\begin{equation}
s_{t} = {\rm Decoder\_T}(s_{t-1}, \hat{y}_{t-1}, \sigma(h_{t}))
\end{equation}
where $\sigma()$ represents a function  to calculate the context vector by using location-sensitive attention mechanism.

Similarly, the student decoder \textit {Decoder\_S} processes the same input sequence and generates the hidden state $\hat{s}_{t}$ at each step $t$ at the same time:

\begin{equation}
\hat{s}_{t} = {\rm Decoder\_S}(\hat{s}_{t-1}, \hat{y}_{t-1}, \sigma(h_{t}))
\end{equation}

In both the teacher model and the student model, the feature loss function $Loss_f$ ensures that the generated speech is close the the target speech,

\vspace{-1em}
\begin{equation}
{Loss_f} = \sum_{t=1}^{T'} L_{2} (\hat{y}_{t},y_{t})
\end{equation}

In the student model, to minimize the discrepancy between the hidden states $s$ and $\hat{s}$ of the teacher model and the student model, we introduce the distillation loss $Loss_d$,

\begin{equation}
{Loss_d} = \frac{1}{T} \sum_{t=1}^{T}|s-\hat{s}|^{2}
\end{equation}

Then the total loss function for the student model is therefore, 
\begin{equation}
{Loss_{total} = Loss_f + \lambda \cdot Loss_d}
\vspace{-3mm}
\label{eqt:loss}
\end{equation}
where $\lambda$ is a trade-off parameter for the two loss terms.

With knowledge distillation, the proposed 2-step teacher-student training scheme allows for a more compact End-to-End network than others such as generative adversarial network\cite{guohh2019gantts}. The teacher model is trained with the objective function $Loss_f$ under the teacher forcing mode, while the student model is trained with a combination of two loss functions $Loss_{total}$ under the free running mode. 

\vspace{-1mm}
\section{Experiments}

We develop two systems on Chinese (12 hours of Data Baker \footnote{https://www.data\--baker.com/open\_source.html}) and English (LJSpeech \footnote{https://keithito.com/LJ-Speech-Dataset/}) corpora separately. To verify the effectiveness of knowledge distillation, denoted as \textit{Tacotron2-KD}, we choose 2 baseline frameworks: 1) Tacotron2 with scheduled sampling, denoted as \textit{Tacotron2-SS}, and  2) Tacotron2 with free running mode, denoted as \textit{Tacotron2-FR}. In all experiments, we use Griffin-Lim algorithm \cite{griffin1984signal} for waveform generation for rapid turn-around.

\begin{table}
\vspace{-5mm}
\centering
 \begin{tabular}{|c|c|c|c|}
 \hline
 \textbf{Framework}& \textbf{Language} &\textbf{MOS}&\textbf{WER}\\
\hline 
\multirow{2}{*}{Tacotron2-SS}  & en & 3.21& 23.82\%\\ 
 & cn & 3.18& 9.44\% \\
\hline
\multirow{2}{*}{\textbf{Tacotron2-KD}}  &  en  & \textbf{3.93}& \textbf{2.17\% }\\ 
 &  cn & \textbf{3.94}& \textbf{0.67\% } \\
\hline
\end{tabular}
\vspace{-3.5mm}
\caption {Comparison of mean opinion scores (MOS) and Word Error Rate (WER\%) between Tacotron2-SS and the proposed Tacotron2-KD.}
\vspace{-5mm}
\end{table}
\vspace{-3.3mm}
\subsection{Experimental Setup}
\vspace{-2mm}
For Chinese experiments, the encoder takes pinyin sequence with tones as input and generates an 160-channel Mel spectrum, two frames at a time, as output. For English experiments, the encoder takes the character sequence as input and generates an 80-channel Mel spectrum, two frames at a time, as output. The two type of encoder inputs are collectively referred to as \textit{character} in this paper. For both systems, we use the Adam optimizer with $\beta_1$ = 0.9, $\beta_2$ = 0.999 and a learning rate of $10^{-3}$ exponentially decaying to $10^{-5}$ starting after 50k iterations. We also apply $L_{2}$ regularization with weight $10^{-6}$. Hyper-parameter $\lambda$ in Equation \ref{eqt:loss} is set as 1.0 and all the models are trained with a batch size of 32. In teacher-student model training, we adopt the teacher model trained with 150k steps as the teacher decoder ``${\rm Decoder\_T}$'', and train the student decoder ``${\rm Decoder\_S}$'' for 150k steps with the proposed knowledge distillation method. 
\vspace{-5mm}
\subsection{Subjective Evaluation}
\vspace{-2mm}
We conduct experiments with out-of-domain test data for naturalness and robustness evaluation.  For Chinese, we select 500 test samples from the Blizzard Challenge 2019 Chinese dataset \cite{cmu-bc2019}. For English, we select 50 test samples from FastSpeech \cite{ren2019fastspeech}, which are particularly hard for TTS system. In addition to the 50 test samples, that are single letters, spellings, repeated numbers, we also include 30 long sentences, each having 128 characters on average. 20 English speakers and 15 Chinese speakers participated in the listening tests.  Each subject listens to 80 converted utterances of his/her native language.
\vspace{-5mm}
\subsubsection{Naturalness Evaluation}
\vspace{-2mm}
We first evaluate the sound quality of the synthesized speech with mean opinion score  (MOS) among Tacotron2-SS, Tacotron2-FR and the proposed Tacotron2-KD, that is reported in Table 1.  The listeners rate the quality on a 5-point scale: “5” for excellent, “4” for good, “3” for fair, “2” for poor, and “1” for bad. It is observed that the proposed Tacotron2-KD clearly outperforms the baseline Tacotron2-SS for both English and Chinese data. As we observe that Tacotron2-FR achieves MOS of 1.33 for English and 2.32 for Chinese, that is significantly lower than those of Tacotron2-SS, we exclude Tacotron2-FR in AB preference test. 

\begin{figure}[tb]
\vspace{-5mm}
\centering
\centerline{\includegraphics[width=8cm]{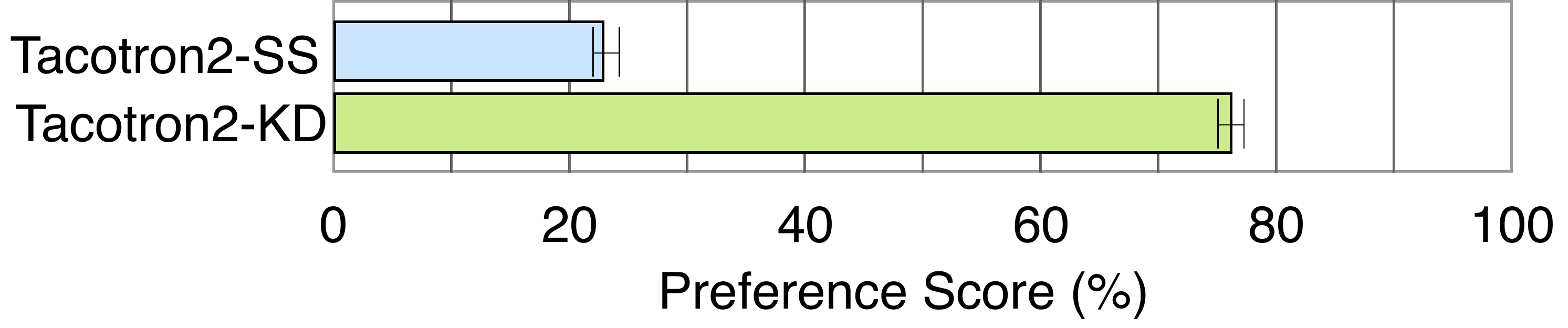}}
\vspace{-4mm}
\caption{The preference test between Tacotron2-KD and Tacotron2-SS on English data, with 95\% confidence interval. }
\label{fig:abtest1}
\end{figure}

\begin{figure}[tb]
\vspace{-2mm}
\centering
\centerline{\includegraphics[width=8cm]{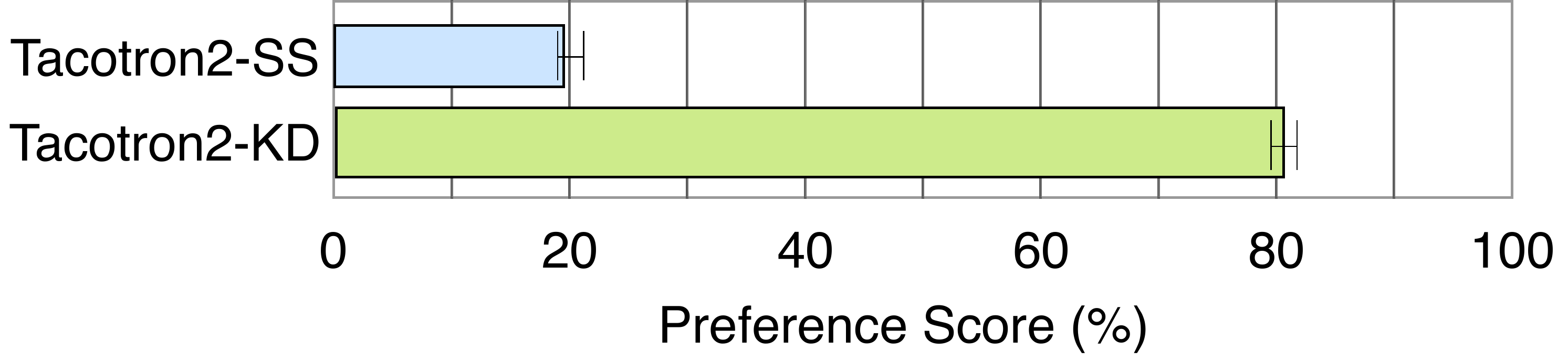}}
\vspace{-4mm}
\caption{The preference test between Tacotron2-KD and Tacotron2-SS on Chinese data, with 95\% confidence interval.}
\vspace{-1em}
\label{fig:abtest2}
\end{figure}

The AB preference test is reported in Figures \ref{fig:abtest1} and \ref{fig:abtest2}, to compare Tacotron2-KD and Tacotron2-SS, in terms of voice quality. It is observed that Tacotron2-KD outperforms Tacotron2-SS consistently for both English and Chinese data.

\vspace{-2em}
\subsubsection{Robustness Evaluation}
\vspace{-2mm}
We further conduct experiments to evaluate the robustness of synthesized speech for Tacotron2-SS and the proposed Tacotron2-KD, as reported in Table 1. We measure the robustness by Word Error Rate (WER \%), that reports the sum of repeats (insertions) and skips (deletions) over the total number of characters in the listening tests \cite{guohh2019gantts}. Repeats and skips represent the two types of errors that Tacotron2 faces. It is shown that Tacotron2-KD effectively reduces the errors by 8.77\% and 21.65\% over the Tacotron2-SS baseline.

A detailed analysis finds that Tacotron2-SS generates 528 skips and 9 repeats for Chinese data, and 251 skips and 12 repeats for English data, while Tacotron2-KD generates only 24 skips for English data and 38 skips for Chinese data. We don't observe any repeats from the Tacotron2-KD outputs, that we think is remarkable.

\vspace{-2mm}
\section{Conclusion}
\vspace{-1mm}
We have studied a training scheme for Tacotron2 to perform high-quality speech synthesis for out-of-domain text,  that overcomes the exposure bias problem. We implement the teacher-student training scheme through a knowledge distillation objective function. We have conducted a series of experiments on both Chinese and English to evaluate the naturalness and robustness. The proposed Tacotron2-KD framework consistently outperforms the baseline systems in both languages.

In addition to the naturalness and robustness improvement, we also discover that Tacotron2-KD delivers improved prosody renderings especially. We will report the prosody analysis of Tacotron2-KD system in the future.

\vspace{-2mm}
\section{Acknowledgements}
\vspace{-2mm}
This research was supported by Human-Robot Interaction Phase 1 (Grant No. 192 25 00054) project, National Research Foundation, Prime Minister’s Office, Singapore under the National Robotics Programme, and by the National Natural Science Foundation of China (No.61563040, No.61773224), Natural Science Foundation of Inner Mongolian (No.2018MS06006, No.2016ZD06) and China Scholarship Council.






\bibliographystyle{IEEEbib}
{\footnotesize
\bibliography{strings,refs}}
\end{document}